\documentclass[runningheads]{llncs}
\usepackage{amsfonts}
\usepackage{amsmath}
\usepackage[T1]{fontenc}
\usepackage{graphicx,verbatim}
\begin{document}
\title{Disentangled and Interpretable Multimodal Attention Fusion for Cancer Survival Prediction}
\titlerunning{Disentangled and Interpretable Multimodal Attention Fusion}
%
\author{Aniek Eijpe\inst{1}\orcidID{0009-0009-7785-8885} \and
Soufyan Lakbir\inst{1}\orcidID{0000-0002-8521-4408} \and
Melis Erdal Cesur\inst{2}\orcidID{0000-0001-8841-1768} \and
Sara P. Oliveira\inst{2}\orcidID{0000-0002-6586-9079} \and
Sanne Abeln\inst{1}\orcidID{0000-0002-2779-7174} \and
Wilson Silva\inst{1}\orcidID{0000-0002-4080-9328}
}
\authorrunning{F. Author et al.}
%
\institute{AI Technology for Life, Department of Information and Computing
Sciences, Department of Biology, Utrecht University, Utrecht, The Netherlands \\ \email{a.eijpe@uu.nl} \and
Computational Pathology group, Department of Pathology, The Netherlands Cancer Institute, Amsterdam, The Netherlands
}


\authorrunning{Eijpe et al.}

\maketitle              
\begin{abstract}

To improve the prediction of cancer survival using whole-slide images and transcriptomics data, it is crucial to capture both modality-shared and modality-specific information.
However, multimodal frameworks often entangle these representations, limiting interpretability and potentially suppressing discriminative features. 
To address this, we propose Disentangled and Interpretable Multimodal Attention Fusion (DIMAF), a multimodal framework that separates the intra- and inter-modal interactions within an attention-based fusion mechanism to learn distinct modality-specific and modality-shared representations. We introduce a loss based on Distance Correlation to promote disentanglement between these representations and integrate Shapley additive explanations to assess their relative contributions to survival prediction. We evaluate DIMAF on four public cancer survival datasets, achieving a relative average improvement of $1.85\%$ in performance and $23.7\%$ in disentanglement compared to current state-of-the-art multimodal models. Beyond improved performance, our interpretable framework enables a deeper exploration of the underlying interactions between and within modalities in cancer biology.


\end{abstract}

\section{Introduction}
Integrating multimodal cancer data offers a promising approach for enhancing deep learning-based survival prediction and personalized cancer care. 
Each modality can provide complementary insights into tumor biology, and a single data modality might therefore not be sufficient to fully capture the heterogeneity and complexity of cancer~\cite{steyaert2023multimodal}. For example, genomics can reveal genetic mutations driving tumor development, while medical images can capture tissue alterations.
 By combining modalities, we can leverage the strengths of each data source to improve predictive accuracy~\cite{boehm2022harnessing,steyaert2023multimodal}, aligning with clinical practice.

Among multimodal strategies, integrating omics data with Whole Slide Images (WSIs) has gained increasing attention~\cite{mobadersany2018predicting,chen2020pathomic,mcat,survpath}. 
Early multimodal approaches typically opted for simple mechanisms, like concatenation~\cite{mobadersany2018predicting}, or a gated Kronecker product~\cite{chen2020pathomic}.
However, these mechanisms generally fail to capture complex interactions between omics data and WSIs~\cite{mcat}. 
Recent advances address this through multilevel fusion, as seen in HBFSurv~\cite{li2022hfbsurv}, or attention-based methods~\cite{mcat,motcat}.
Among the latest attention-based frameworks, SurvPath~\cite{survpath} processes bulk transcriptomics data with biological pathways~\cite{hallmark} to generate semantically meaningful representations and fuses them with WSI patch features using a memory-efficient transformer.
Building on this, MMP~\cite{mmp} summarizes the WSIs with Gaussian Mixture Models (GMMs)~\cite{panther}, paving the way for full transformer-based fusion and post-hoc interpretability analysis.

Despite significant advances, multimodal methods focus primarily on shared information across modalities~\cite{survpath,mmp,motcat,mcat}, leading to the suppression of modality-specific information~\cite{pibd}.
While shared information can enhance robustness~\cite{steyaert2023multimodal}, information specific to each modality can provide additional discriminative power.
To address this, Disentangled Representation Learning (DRL)~\cite{bengio2013representation} can be used to decompose the input into modality-shared and modality-specific representations.
Recent works have explored this, promoting disentanglement with minimizing an orthogonal~\cite{wu2023camr} or a contrastive loss~\cite{robinet2024drim}, and using adversarial training~\cite{wu2023camr,robinet2024drim}. 
Similarly, PIBD~\cite{pibd} promotes disentanglement by minimizing a contrastive log-ratio upper bound of mutual information~\cite{cheng2020club}. 
However, none of these methods have explicitly measured disentanglement or assessed the contribution of the modality-specific and modality-shared representations to survival prediction.

In this work, we introduce \textbf{Disentangled and Interpretable Multimodal Attention Fusion (DIMAF)}, an inherently interpretable framework that disentangles the fusion between WSIs and bulk transcriptomics for cancer survival prediction.
Specifically, we use two self-attention layers to capture the intra-modal interactions and two cross-attention layers to model the inter-modal interactions, resulting in separate modality-specific and modality-shared representations. Unlike MMP~\cite{mmp}, we disentangle the multimodal interactions to provide a structured, disentangled view of multimodal information, enhancing interpretability and eliminating the need for post-attention FNNs.
Measuring and enforcing disentanglement between latent representations without the ground truth is challenging and remains underexplored~\cite{carbonneau2022measuring}. To address this, we employ Distance Correlation (DC)~\cite{empericaldcor,liu2020metrics}, a statistical metric measuring the dependence between two representations. DC does not require pre-defined kernels, as the Hilbert-Schmidt independence criterion~\cite{gretton2005measuring} does, and captures both linear and non-linear dependencies, unlike orthogonality constraints.
We assess the interpretability of our framework by having a pathologist evaluate the learned WSI representations and use SHapley Additive exPlanations (SHAP)~\cite{NIPS2017_7062}, a feature attribution method, to quantify the relative contributions of the modality-specific and modality-shared representations.

We evaluate DIMAF on four public cancer survival datasets, demonstrating improved performance and disentanglement over state-of-the-art multimodal models. Beyond improved performance, our framework provides a strong foundation for deeper exploration of multimodal cancer biology.
In summary, our work makes the following key contributions:
\begin{itemize}
    \item We introduce \textbf{DIMAF}, an interpretable multimodal cancer survival model that explicitly disentangles the intra- and inter-modal interactions between WSIs and transcriptomics data within an attention-based fusion mechanism.
    \item To promote disentanglement and learn distinct modality-specific and modality-shared representations, we incorporate a disentanglement loss based on DC.
    \item We employ SHAP to gain insights into the importance of modality-shared and modality-specific representations.
    \item We evaluate DIMAF on four public cancer survival datasets, demonstrating enhanced performance, disentanglement, and interpretability.
\end{itemize}

\section{Methodology}

\begin{figure}[t]
\centering
\includegraphics[width=\textwidth]{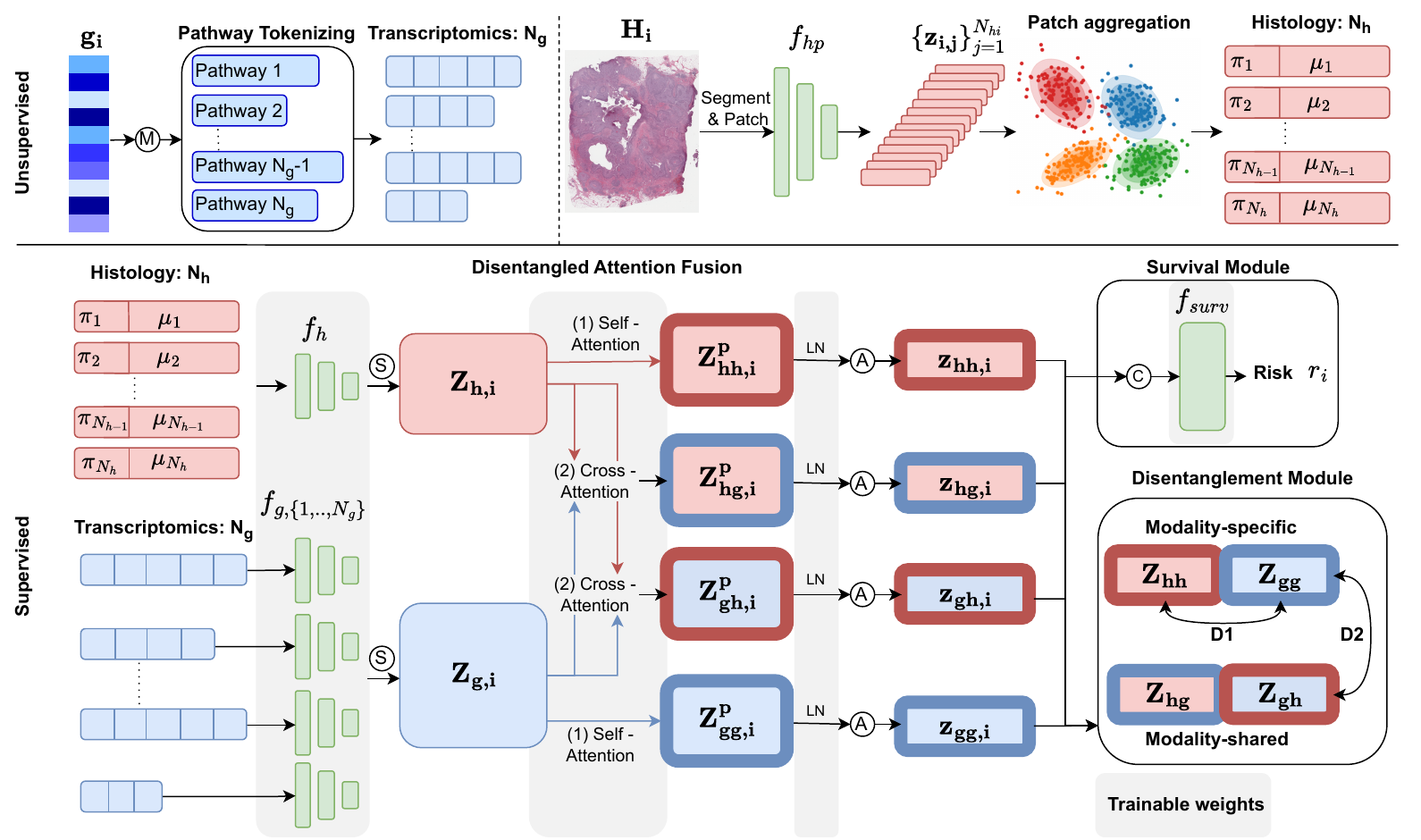}
\caption{Overview of the proposed framework. First, the transcriptomics data $\mathbf{g_i}$ is processed through pathway tokenizing and $f_{g,\{1,..,N_g\}}$ to obtain the final pathway embedding $\mathbf{Z_{g,i}}$. From the WSI $\mathbf{H_i}$, we obtain patch features $\{ \mathbf{z_{i,j}} \} _{j=1}^{N_{hi}}$, which are aggregated and processed through $f_h$ to obtain the slide embedding $\mathbf{Z_{h,i}}$. Then, $\mathbf{Z_{g,i}}$ and $\mathbf{Z_{h,i}}$ are processed through the Disentangled Attention Fusion module, obtaining $\mathbf{z_{hh, i}}, \mathbf{z_{hg, i}}, \mathbf{z_{gh, i}}, \text{ and }\mathbf{z_{gg, i}}$. These representations are further disentangled in the Disentanglement Module and used to compute the survival risk $r_i$ in the Survival Module.} \label{fig:framework}
\end{figure}

We propose a disentangled approach for survival analysis that leverages multimodal data. Specifically, for each patient $i$, the objective is to obtain a risk prediction $r_i$, given their WSI $\mathbf{H_i}$ and transcriptomics data $\mathbf{g_i}$. Figure~\ref{fig:framework} provides an overview of our framework, consisting of three main components: unimodal feature extraction, disentangled attention fusion, and survival prediction.
\subsection{Unimodal feature extraction}\label{sec:unifeat}
We process the transcriptomics data and WSIs separately to obtain interpretable unimodal representations. Following prior methods~\cite{pibd,mmp}, we tokenize the transcriptomics data $\mathbf{g_i} \in \mathbb{R}^{D_g}$, representing expression levels of $D_g$ different genes, by using $N_g$ biological pathways as interpretable features. Each pathway represents a group of genes active in specific biological processes in cancer~\cite{hallmark}. The tokenized gene expressions are processed through separate Self-Normalizing Networks (SNNs)~\cite{snn} ($f_{g,\{1,..,N_g\}}$) and $N_g$ random-initialized and learnable pathway encodings~\cite{mmp} are concatenated to each of the obtained pathway embeddings. Lastly, these pathway embeddings with the concatenated encodings are stacked (S) to obtain the final pathway embedding $\mathbf{Z_{g, i}} \in \mathbb{R}^{N_g\times D_{gh}}$. 

The WSI $\mathbf{H_i}$ is split into $N_{hi}$ non-overlapping patches, from which features are extracted using a frozen encoder $f_{hp}$. To obtain a slide representation from these patch features, we use PANTHER~\cite{panther,mmp}. Specifically, all the patch features $\{\mathbf{z_{i, j}}\}_{j=1}^{N_{hi}}$ are fitted to a GMM, where $\{\pi_{c}, \boldsymbol{\mu_c}, \Sigma_c\}_{c=1}^{N_h}$ denote the learned importance, mean, and variance of each Gaussian component $c$, respectively.
The importance $\pi_c$ and mean $\boldsymbol{\mu_c}$ of each component $c$ are concatenated, passed through an MLP, concatenated with a learnable prototype encoding and stacked to obtain the WSI embedding $\mathbf{Z_{h,i}} \in \mathbb{R}^{N_h\times D_{gh}}$. Intuitively, the mean $\boldsymbol{\mu_c}$ of each Gaussian represents a morphological prototype, while the importance $\pi_c$ reflects the proportion of patches in the WSI associated with that prototype. To interpret the prototypes, we can visualize the patches $\mathbf{z_{i, j}}$ for which component $c$ has the highest probability, determined by the posterior distribution $q(c|\mathbf{z_{i,j}})$~\cite{panther,mmp}.

\subsection{Disentangled attention fusion}
\noindent With the obtained unimodal representations $\mathbf{Z_{h,i}}$ and $\mathbf{Z_{g,i}}$, we perform multimodal fusion using a disentangled attention mechanism. Within this mechanism, we distinctly model the intra-modal interactions by using two self-attention layers to encode the modality-specific information of the transcriptomics data and WSI data within $\mathbf{Z_{gg, i}^{p}} \in \mathbb{R}^{N_g\times D_{z}}$ and $\mathbf{Z_{hh, i}^{p}} \in \mathbb{R}^{N_h\times D_{z}}$, respectively (Eq.~\ref{eq:self}).

\begin{equation}\label{eq:self}
        \mathbf{Z_{\{gg, hh\}, i}^{p}} = \sigma \biggl( \frac{(\mathbf{Z_{\{g, h\},i}W^{\{g,h\}}_{Q1}})(\mathbf{Z_{\{g, h\},i}W^{\{g,h\}}_{K1}})^T}{\sqrt{d}}\biggr) (\mathbf{Z_{\{g, h\},i}W^{\{g,h\}}_{V1}}) 
\end{equation}

\noindent  Here, d denotes a scaling parameter stabilizing the attention scores.
We model the inter-modal interactions to encode the modality-shared information within $\mathbf{Z_{hg, i}^{p}} \in \mathbb{R}^{N_g\times D_{z}}$ and $\mathbf{Z_{gh, i}^{p}} \in \mathbb{R}^{N_h\times D_{z}}$, by using two cross-attention layers (Eq.~\ref{eq:cross}). 

\begin{equation}\label{eq:cross}
    \mathbf{Z_{\{gh, hg\}, i}^{p}} = \sigma \biggl( \frac{(\mathbf{Z_{\{h, g\},i}W^{\{h,g\}}_{Q2}})(\mathbf{Z_{\{g, h\},i}W^{\{g,h\}}_{K2}})^T}{\sqrt{d}}\biggr) (\mathbf{Z_{\{g, h\},i}W^{\{g,h\}}_{V2}}) 
\end{equation}

\noindent On the obtained representations, we apply Layer Normalization (LN) and average over the pathways or GMM components per representation, obtaining the final disentangled multimodal representation $[\mathbf{z_{gg, i}}$, $\mathbf{z_{hh, i}}$, $\mathbf{z_{hg, i}}$, $\mathbf{z_{gh, i}}] \in \mathbb{R}^{4 \cdot D_z}$.

By using different learnable Query, Key, and Value weight matrices ($\mathbf{W_Q}$, $\mathbf{W_K}$, $\mathbf{W_V}$) for the different types of connections (Eq.~\ref{eq:self} and~\ref{eq:cross}) and modalities, the model is encouraged to focus on different aspects of the data, facilitating effective disentanglement. However, this capacity alone does not guarantee disentanglement, necessitating an explicit disentanglement loss. We use DC to promote disentanglement between the modality-specific representations (D1 in Figure~\ref{fig:framework}) and between the modality-specific and modality-shared representations (D2).

\begin{equation}\label{eq:dcorloss}
    \mathcal{L}_{dis} = DC(\mathbf{Z_{gg}, Z_{hh}}) + DC([\mathbf{Z_{gg}, Z_{hh}}], [\mathbf{Z_{hg}, Z_{gh}}])
\end{equation}
\begin{equation}\label{eq:dcor}
    \text{DC}(\mathbf{Z_1, Z_2}) = \frac{dCov(\mathbf{Z_1, Z_2})}{\sqrt{dCov(\mathbf{Z_1, Z_1})dCov(\mathbf{Z_2, Z_2})}}
\end{equation}
\noindent where $\mathbf{Z_{gg}, Z_{hh}, Z_{hg}, \text{ and } Z_{gh}} \in \mathbb{R}^{B\times D_z}$ are the $B$ stacked disentangled representations. $dCov(\mathbf{X, Y})$ denotes the (Euclidean) distance covariance between $\mathbf{X}$ and $\mathbf{Y}$~\cite{empericaldcor}. Intuitively, $dCov(\mathbf{X, Y})$ measures dependence by comparing pairwise distances within $\mathbf{X}$ and $\mathbf{Y}$, assessing how variations in one correspond to variations in the other. If two representations are independent, their variations are unrelated, resulting in a small $dCov$. DC normalizes $dCov$ such that $ \text{DC} \in [0,1]$.
\subsection{Survival Analysis}
Finally, for the survival prediction, we utilize the Cox proportional hazards model~\cite{cox1972regression} on the concatenated disentangled representations.
\begin{equation}
    h(t|\mathbf{z_{mm,i}}) = h_0(t) \cdot e^{r_i} = h_0(t) \cdot e^{f_{surv}([\mathbf{z_{gg, i}}, \mathbf{z_{hh, i}}, \mathbf{z_{hg, i}}, \mathbf{z_{gh, i}}])} 
\end{equation}
$h_0(t)$ denotes the baseline hazard, $f_{surv}$ is a linear predictor, and $r_i$ is the risk score of patient $i$, which is optimized with the Cox partial log likelihood~\cite{wong1986theory}. 
\begin{equation}
    \mathcal{L}_{surv} = -\sum_{i \in U}\bigg(r_i - log \big(\sum_{j \in K_i}e^{r_j}\big)\Bigg)
\end{equation}
\noindent Here, $U$ is the set of uncensored patients and $K_i$ is the set of patients whose time of death or last known follow-up time is after $i$.

\section{Experiments}
\subsection{Experimental settings}
\subsubsection{Dataset} We evaluate our approach on four public cancer datasets from The Cancer Genome Atlas\footnote{https://portal.gdc.cancer.gov} 
(TCGA): Breast Invasive Carcinoma (BRCA, $n=868$), Bladder Urothelial Carcinoma (BLCA, $n=359$), Lung Adenocarcinoma (LUAD, $n=412$), and Kidney Renal Clear Cell Carcinoma (KIRC, $n=340$). The transcriptomics data is obtained via the XENA database~\cite{goldman2020visualizing} and includes RNA-seq expression levels that are normalized across all TCGA cohorts. Moreover, Disease-Specific Survival (DSS) is added to the clinical data as survival endpoint~\cite{liu2018integrated}, as a better estimate of disease status than Overall Survival (OS).

\subsubsection{Baselines}
First, we compare our method against multivariate linear Cox regression~\cite{lifelines_coxph} as a clinical baseline using grade, age, and sex (if available). Moreover, we employ three state-of-the-art multimodal cancer survival models using WSIs and transcriptomics data, which demonstrated superior performance over unimodal models.
SurvPath~\cite{survpath} tokenizes the transcriptomics data via biological pathways and approximates transformer-based fusion.
MMP~\cite{mmp} summarizes the WSIs with morphological prototypes for full transformer-based fusion with the same pathway data. PIBD~\cite{pibd} uses prototypical information bottlenecking and tries to disentangle modality-specific and modality-shared representations. 

\subsubsection{Implementation details}
For all multimodal models, WSIs were split into $256\times256$ px patches with a resolution of $0.5 \mu m$ per pixel using CLAM~\cite{clam} and the patch features were extracted with UNI~\cite{uni,oquab2023dinov2}. Additionally, the same $N_g=50$ Hallmark pathways were used~\cite{hallmark}.
The models were trained for 30 epochs with the AdamW optimizer, a learning rate of $1e^{-4}$, a cosine learning scheduler, and $1e^{-5}$ weight decay. A batch size of 64 was used, except for SurvPath, which required a batch size of 1. The number of mixture distributions $N_h$ was set to 16 for both MMP and DIMAF. The total loss function used for optimization is defined as $\mathcal{L} = \lambda_{surv} \cdot \mathcal{L}_{surv} +  \lambda_{dis} \cdot \mathcal{L}_{dis}$, with $\lambda_{\text{surv}} = 1$ and $\lambda_{\text{dis}} = 7$.
We performed 5-fold cross-validation to predict DSS, evaluating model performance with the concordance-index (c-index)~\cite{harrell1996multivariable}.
To mitigate potential batch effects, train-test splits were stratified by sample site~\cite{survpath}. Additionally, the disentanglement in PIBD and DIMAF was assessed using DC.

\subsection{Results}
\subsubsection{Survival analysis}

Table \ref{surv} presents the DSS test results across four cancer types. 
Our proposed DIMAF model achieves the highest average c-index with a relative improvement of $1.85\%$ compared to the best-performing baseline MMP and $12.2\%$ compared to the DRL baseline PIBD. 
The ablation variant DIMAF$_{\neg\text{dis}}$, from which the disentanglement objective (Eq.~\ref{eq:dcorloss}) is removed, performs comparably in terms of survival prediction. Its impact varies by dataset, indicating that disentanglement affects cancer types in distinct ways. Overall, these results highlight DIMAF's improved performance in integrating multimodal data for survival prediction.
\begin{table}[t]
\centering
\caption{DSS test results (c-index) over the different folds (mean $\pm$ std). The best and second-best performances are denoted by \textbf{bold} and \underline{underlined}, respectively.}\label{surv}
\begin{tabular}{|l|l|l|l|l|l|}
\hline
\textbf{Model} & \textbf{BRCA}  & \textbf{BLCA} &  \textbf{LUAD} & \textbf{KIRC} & \textbf{Avg.} ($\uparrow$) \\
\hline
 Clinical & $0.491\pm0.113$ & $0.519\pm0.067$ & $0.461\pm0.057$ & $0.533\pm0.083$ & $0.501$\\
 \hline
 PIBD~\cite{pibd} & $0.671\pm0.036$ & $0.573\pm0.029 $& $0.578\pm0.027$ & $0.726\pm0.114$ & $0.637$ \\
 
 SurvPath~\cite{survpath} & $0.701\pm0.017$ & $0.610\pm0.055$ & $0.611\pm0.060$ & $0.737\pm0.100$ & $0.665$ \\
 
 MMP~\cite{mmp} & $0.750\pm0.052$ &$0.656\pm0.046$ &$ \mathbf{0.674\pm0.038}$ &$ 0.728\pm0.106$ & $0.702$\\

  DIMAF$_{\neg\text{dis}}$& $\mathbf{0.769\pm0.044}$ & $\underline{0.675\pm0.034}$ & $0.651\pm0.043$ & $\underline{0.747\pm0.093}$& $\underline{0.711}$\\
 
 DIMAF & $\underline{0.759\pm0.067}$ & $\mathbf{0.679\pm0.043}$ & $\underline{0.669\pm0.062}$ & $\mathbf{0.752\pm0.092}$ & $\mathbf{0.715}$\\
 
\hline
\end{tabular}
\end{table}

\subsubsection{Disentanglement analysis}
Table~\ref{disentanglement} presents the test results of the disentanglement between the two modality-specific representations (D1 in Figure~\ref{fig:framework}), the disentanglement between the modality-specific and modality-shared representations (D2 in Figure~\ref{fig:framework}), and the total disentanglement (see Eq.~\ref{eq:dcorloss}). DIMAF achieves the lowest average DC for all three types, indicating the highest degree of disentanglement. It substantially outperforms PIBD, the only baseline designed for disentanglement, by a relative average improvement of $23.7\%$. Notably, DIMAF increases D2 disentanglement by $27.7\%$ over PIBD and $31.1\%$ over DIMAF$_{\neg\text{dis}}$, further supporting DIMAF's effectiveness in obtaining disentangled multimodal representations and highlighting the importance of the disentanglement loss.
While some correlation remains, DIMAF shows substantial progress in disentangling modality-specific and modality-shared representations.
\begin{table}[t]
\centering
{\fontsize{8.5}{9.5}\selectfont
\caption{Disentanglement test results (DC) over the different folds (mean $\pm$ std). The best total performances are denoted by \textbf{bold}.}\label{disentanglement}
\begin{tabular}{|l|l|l|l|l|l|l|}
\hline
\textbf{Model} & \textbf{Type} &  \textbf{BRCA} & \textbf{BLCA} &  \textbf{LUAD} & \textbf{KIRC} & \textbf{Avg.} ($\downarrow$) \\
\hline
 PIBD~\cite{pibd} & D1 &$ 0.346\pm0.011$ & $0.454\pm0.033$ & $0.455\pm0.038$ & $0.534\pm0.021$ & $0.447$ \\
                  & D2 & $0.922\pm0.049$ & $0.920\pm0.054$ & $0.957\pm0.025$ & $0.908\pm0.019$ & $0.927$ \\
                  & Total & $0.634\pm0.026$ & $0.687\pm0.025$ & $0.706\pm0.028$ & $\mathbf{0.721\pm0.007}$ & $0.687$ \\
 \hline
 DIMAF$_{\neg\text{dis}}$& D1 & $0.505\pm0.018$ & $0.641\pm0.04$ & $0.602\pm0.07$ & $0.600\pm0.084$ & $0.587$ \\
   & D2  & $0.952\pm0.017$ & $0.977\pm0.003$ & $0.973\pm0.005$ &$ 0.985\pm0.006$ & $0.972$ \\
 & Total  & $0.728\pm0.015$ & $0.809\pm0.02$ & $0.787\pm0.036$ & $0.792\pm0.043$ & $0.779$ \\
 \hline
 DIMAF & D1 & $0.245\pm0.042$ & $0.401\pm0.029 $& $0.339\pm0.046$ & $0.525\pm0.089$ & $0.378$ \\
  & D2 & $0.468\pm0.049$ & $0.672\pm0.068$ & $0.609\pm0.029$ & $0.931\pm0.031$ & $0.670$ \\
 & Total & $\mathbf{0.356\pm0.044}$ & $\mathbf{0.537\pm0.048}$ & $\mathbf{0.474\pm0.018}$ & $0.728\pm0.053$ & $\mathbf{0.524}$ \\	
 
\hline
\end{tabular}
}
\end{table}

\subsubsection{Interpretability}
We evaluated the BRCA slide representations by analyzing the morphological patterns captured by the learned prototypes (Section~\ref{sec:unifeat}). 
A pathologist annotated the patch clusters, confirming that the prototypes capture distinct morphological structures, including tumor and immune cells, adipose tissue, and stroma, with some overlap within and between prototypes. Furthermore, the structures seen in the clusters were consistent across train and test sets. 
\begin{table}[t]
\centering
\caption{Normalized SHAP values (\%) of the disentangled representations (\textbf{Repr.}) over the different folds (mean $\pm$ std).}\label{shap}
\begin{tabular}{|l|l|l|l|l|l|l|}
\hline
\textbf{Model} & \textbf{Repr.} &   \textbf{BRCA} & \textbf{BLCA} &  \textbf{LUAD }& \textbf{KIRC} & \textbf{Avg.}\\
\hline
 DIMAF$_{\neg\text{dis}}$& Specific & $0.513\pm0.033$ & $0.548\pm0.028$ & $0.538\pm0.028$ & $0.517\pm0.024$ & $0.529$ \\
  & Shared & $0.487\pm0.033$ & $0.452\pm0.028$ & $0.462\pm0.028$ & $0.483\pm0.024$ & $0.471$\\
  \cline{2-7}
  & $\mathbf{Z^p_{hh}}$ & $0.315\pm0.024$ & $0.325\pm0.047$ & $0.291\pm0.034$ & $0.294\pm0.027$ & $0.306$ \\
  & $\mathbf{Z^p_{gg}}$ & $0.207\pm0.040$ & $0.223\pm0.028$ & $0.237\pm0.017$ & $0.232\pm0.032$ & $0.225$ \\
  & $\mathbf{Z^p_{gh}}$ & $0.188\pm0.022$ & $0.209\pm0.033$ & $0.218\pm0.022$ & $0.201\pm0.032$ & $0.204$ \\
  & $\mathbf{Z^p_{hg}}$ & $0.290\pm0.043$ & $0.243\pm0.020$ & $0.253\pm0.027$ &  $0.273\pm0.043$ & $0.265$ \\

 \hline
  DIMAF & Specific & $0.180\pm0.037$ & $0.273\pm0.038$ & $0.257\pm0.027$ & $0.302\pm0.043$ & $0.253$ \\
 & Shared & $0.820\pm0.037$ & $0.727\pm0.038$ & $0.743\pm0.027$ & $0.698\pm0.043$ & $0.747$ \\
  \cline{2-7}
   & $\mathbf{Z^p_{hh}}$ & $0.083\pm0.03$ & $0.108\pm0.022$ & $0.071\pm0.015$ & $0.073\pm0.033$ & $0.084$ \\
  & $\mathbf{Z^p_{gg}}$ & $0.111\pm0.024$ & $0.176\pm0.028$ & $0.183\pm0.023$ & $0.232\pm0.031$ & $0.176$ \\
  & $\mathbf{Z^p_{gh}}$ & $0.300\pm0.046$ & $0.322\pm0.042$ & $0.357\pm0.023$ & $0.298\pm0.039$ & $0.319$ \\
  & $\mathbf{Z^p_{hg}}$ & $0.506\pm0.078$ & $0.394\pm0.064$ & $0.389\pm0.026$ & $0.397\pm0.092$ & $0.421$ \\

\hline
\end{tabular}
\end{table}

Next, we investigated the relative contributions of the modality-specific ($\mathbf{Z_{hh}^{p}}$ and $\mathbf{Z_{gg}^{p}}$) and modality-shared ($\mathbf{Z_{gh}^{p}}$ and $\mathbf{Z_{hg}^{p}}$) representations obtained by Deep SHAP~\cite{NIPS2017_7062} over all four datasets (see Table~\ref{shap}). Without the disentanglement loss (DIMAF$_{\neg\text{dis}}$), the model relies almost equally on the shared and specific representations. In contrast, DIMAF shifts the focus towards the shared representations, suggesting that many discriminative features are inherently shared but are incorrectly captured in modality-specific representations without explicit disentanglement. However, the observed contribution of the modality-specific representations highlights the importance of explicitly modeling these features to preserve important information.
Notably, the contribution of the WSI-specific representation ($\mathbf{Z_{hh}^{p}}$) decreases by $72.5\%$ when disentanglement is promoted, suggesting that without disentanglement, $\mathbf{Z_{hh}^{p}}$ contains a substantial amount of shared features. The contribution of the transcriptomics-specific representation ($\mathbf{Z_{gg}^{p}}$) also decreases with disentanglement, but only by $21.8\%$. 
It is important to note here that the dimensions of $\mathbf{Z^p_{gg}}$ and $\mathbf{Z^p_{hh}}$ differ, as the number of pathways does not match the number of GMM elements (i.e., $N_g \neq N_h$). Future analyses into the SHAP values of the various morphological prototypes and pathways could offer deeper insights into the contributions of modality-specific features. 
Additionally, when comparing with Table~\ref{disentanglement}, a potential correlation emerges between the degree of disentanglement and the shift in SHAP values, i.e., higher disentanglement appears to increase the SHAP values for shared representations. 
Overall, DIMAF's disentanglement strategy highlights shared representations as the most informative for survival prediction. However, the presence of important features in the specific representations shows the value of explicitly modeling the modality-specific information.

\section{Conclusion}
In this work, we introduced \textbf{DIMAF}, an interpretable framework for cancer survival prediction that explicitly disentangles intra- and inter-modal interactions within the fusion of WSIs and transcriptomics data. 
We demonstrated that DIMAF improves state-of-the-art performance and disentanglement across four public cancer survival datasets. The SHAP-based interpretability analysis revealed the importance of modality-shared representations while confirming that modality-specific features still contribute to survival prediction.

Our framework provides a strong foundation for further interpretability analysis, enabling a more comprehensive understanding of multimodal cancer biology.
Future work can explore the inherent interpretability of DIMAF by combining the attention weights of the intra- and inter-modal interactions with SHAP values before and after fusion to capture both feature importance and multimodal interactions. This dual approach will offer more robust insights into how different data modalities interact in DIMAF for accurate survival prediction.

%
%
%
\bibliographystyle{splncs04}
\bibliography{mybib}
\end{document}